\documentclass[10pt,journal,compsoc]{IEEEtran}
\ifCLASSOPTIONcompsoc
  \usepackage[nocompress]{cite}
\else
  \usepackage{cite}
\fi
\usepackage{color}
\usepackage{times}
\usepackage{epsfig}
\usepackage{amsmath}
\usepackage{amssymb}
\usepackage{caption}
\usepackage{multirow}
\usepackage{graphicx}
\newcommand{\etal}{\textit{et al}.}
\usepackage[colorlinks,linkcolor=red]{hyperref}

\usepackage{wrapfig}
\usepackage{hyphenat}
\usepackage{soul,color}
\usepackage{amsopn}
\usepackage{amsmath}
\usepackage{amssymb}
\usepackage{tabularx}
\usepackage{subfigure}

\DeclareMathOperator*{\argmin}{argmin}

\definecolor{Red}{cmyk}{0,1,1,0}
\definecolor{Green}{cmyk}{1,0,1,0}
\definecolor{Cyan}{cmyk}{1,0,0,0}
\definecolor{Purple}{cmyk}{0.45,0.86,0,0}
\definecolor{Rosolic}{cmyk}{0.00,1.00,0.50,0}
\definecolor{Blue}{cmyk}{1.00,1.00,0.00,0}

\newcommand{\youyi}[1]{{\color{black} #1}}

\ifCLASSINFOpdf
\else
\fi
\begin{document}
%
\title{\emph{AutoSweep}: Recovering 3D Editable Objects from a Single Photograph}
%
%
%
%

\author{Xin~Chen\quad
        Yuwei~Li \quad
        Xi Luo \quad
        Tianjia Shao \quad
        Jingyi Yu \quad
        Kun Zhou \quad
        Youyi Zheng$^{\dag}$
        \thanks{$\dag$ corresponding author}
\IEEEcompsocitemizethanks{\IEEEcompsocthanksitem X. Chen, Y. Li, X. Luo, and J. Yu are with School of Information Science and Technology,ShanghaiTech University, China.\protect\\
E-mail: ~\{chenxin2,liyw,luoxi,yujingyi\}@shanghaitech.edu.cn

\IEEEcompsocthanksitem T. Shao is with the School of Computing, University of Leeds., UK.\protect\\
E-mail: ~T.Shao@leeds.ac.uk

\IEEEcompsocthanksitem Y. Zheng, and K. Zhou are with the State Key Lab of CAD\&CG, Zhejiang University, China.\protect\\
E-mail: ~zyy, kunzhou@cad.zju.edu.cn

}
}

%
%

\markboth{Journal of \LaTeX\ Class Files,~Vol.~14, No.~8, August~2015}%
{Shell \MakeLowercase{\textit{et al.}}: Bare Demo of IEEEtran.cls for Computer Society Journals}
%



\IEEEtitleabstractindextext{%
\begin{abstract}
This paper presents a fully automatic framework for extracting editable 3D objects directly from a single photograph. Unlike previous methods which recover either depth maps, point clouds, or mesh surfaces, we aim to recover 3D objects with semantic parts and can be directly edited. We base our work on the assumption that most human-made objects are constituted by parts and these parts can be well represented by generalized primitives. Our work makes an attempt towards recovering two types of primitive-shaped objects, namely, generalized cuboids and generalized cylinders. To this end, we build a novel instance-aware segmentation network for accurate part separation. Our GeoNet outputs a set of smooth part-level masks labeled as profiles and bodies. Then in a key stage, we simultaneously identify profile-body relations and recover 3D parts by sweeping the recognized profile along their body contour and jointly optimize the geometry to align with the recovered masks. Qualitative and quantitative experiments show that our algorithm can recover high quality 3D models and outperforms existing methods in both instance segmentation and 3D reconstruction. The dataset and code of AutoSweep are available at \href{https://chenxin.tech/AutoSweep.html}{https://chenxin.tech/AutoSweep.html}.

\end{abstract}

}

\maketitle

\IEEEdisplaynontitleabstractindextext

%
\IEEEpeerreviewmaketitle

\IEEEraisesectionheading{\section{Introduction}\label{sec:introduction}}

\IEEEPARstart{T}{here} is an emerging demand on automatic extraction of high quality 3D objects from a single photograph. Applications are numerous, ranging from image manipulation \cite{Zhou:2010:PRH,Zheng:2012:IIC,kholgade20143d}, to emerging 3D printing \cite{Prevost:2013:MSB,Dumas:2014:BGA} and virtual reality and augmented reality \cite{Apple:Ar, Arora:vrSketching:2017}. For example, in e-commerce, it is highly desirable to automatically and quickly recover the 3D model of a commercial product from its 2D image (e.g., in advertisement). Further, the geometry and the texture map should be of high quality to be useful. The problem, however, remains challenging: any successful solution should be able to reliably segment an object from the image and then recover its shape and structure whereas both problems are ill-posed and generally require imposing priors and using sophisticated optimization.

A photograph is inherently ``flat" and does not contain associated depth information. Traditional solutions rely on multi-view stereo or volumetric reconstructions to recover the point cloud, normal, or visual hull of the object. They require using multiple images of an object which is most likely inaccessible in applications such as e-commerce. More importantly, the recovered 3D geometry is of low quality even with the most advanced reconstruction algorithms. Alternative solutions \cite{chen20133,Debevec:CSD96,Jiang:2009:SAM,Zheng:2012:IIC} treat an object as a composition of simple, primitive components \cite{Brooks:1979,GuptaEfrosHebert_ECCV10} and set out to estimate each individual component. Most existing methods in this category require extensive human inputs for partitioning the object. Most recently, end-to-end methods \cite{choy20163d,fan2016point} have leveraged generative neural networks to directly infer point cloud or volumetric representations of an object from a single image. They are able to produce coarse geometry that resembles the actual shape. Yet, the quality of the resulting model still barely meet the one of a CAD model or a parametric mesh.

\begin{figure}[b!]
    \centering
    \includegraphics[width=\linewidth]{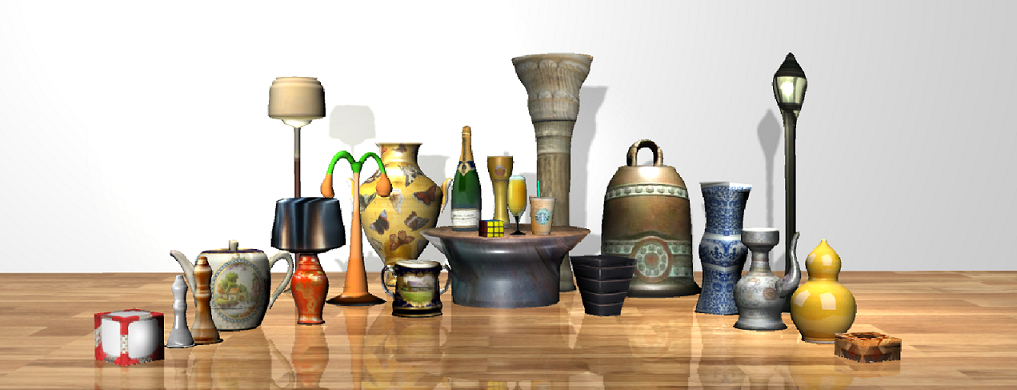}
    \caption{Exemplar 3D models generated using our method.}
    \label{fig:teaser}
\end{figure}

\begin{figure*}[t!]
  \centering
  \includegraphics[width=\linewidth]{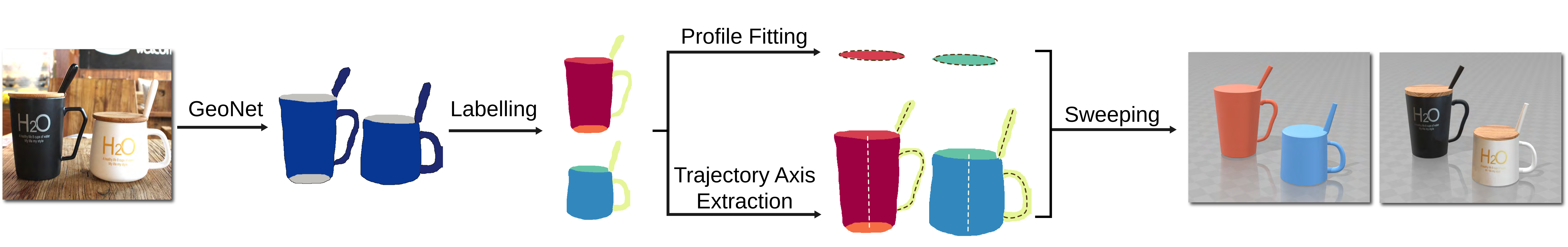}
  \caption{The pipeline. Our method takes as input a single photograph and extracts its semantic part masks labeled as \emph{cylinder profile}, \emph{cuboid profile}, \emph{cylinder body}, etc., which are then used in a sweeping procedure to construct a textured 3D model.}\label{fig:modeling-pipeline}
\end{figure*}

In this paper, we present a fully automatic, single-image based technique for producing very high quality 3D geometry of a specific class of objects: objects composed of generalized cuboids and generalized cylinders or \textit{GC-GCs}, for short. \textcolor[rgb]{0.00,0.00,0.00}{Both a generalized cuboid and a generalized cylinder could be represented as a profile (i.e., a circle or a rectangle) sweeping along a trajectory axis as in traditional CAD systems. Normally, the profile is allowed to scale and the trajectory axis is curved \cite{chen20133}.}
An intriguing benefit of our reconstruction pipeline is that each cuboid and cylindrical part can be directly edited by altering the profile or the trajectory axis and then composed together to form a new GC-GC. See Fig. \ref{fig:teaser}, \ref{fig:synthdata} for examples of GC-GCs. 

In our solution, we first partition and recognize each semantic part of a GC-GC object. We exploit instance segmentation network Mask R-CNN \cite{he2017mask} which is capable of handling ``invisible profiles" that caused by occlusions of the foreground or even self occlusions. However, due to a small receptive field, the output often contains erroneous boundaries and incomplete masks that do not agree with the actual object mask. We extend the structure of Mask R-CNN and construct our Geometry Network (\emph{GeoNet} for short) by incorporating contour and edge maps into a concatenating network which we call the deformable convolutional network (DCN) derived from \cite{chen2016deeplab,dai2017deformable}. The edge maps and 2D contours are used to better learn the boundary of the body and face regions which are crucial in the subsequent modeling process. Our network outputs smooth masks around the boundary regions.


Once we segment each component, we conduct reconstruction via a volume sweeping scheme. We decouple the process into two stages of profile fitting and profile sweeping. To estimate the 3D profile, we jointly optimize the profile with the camera pose. We then extract the trajectory axis of each body mask and map it to 3D with the estimated camera pose to guide the optimization of the profile sweeping.

We demonstrate our approach to various images. Our system is capable of automatically generating 3D models from a single photograph which can then be used for editing and rearranging. Qualitative and quantitative experiments are conducted to verify the effectiveness of our method.

\section{Related Work}
\textbf{Semantic Segmentation.} Recent deep neural networks have shown great success in improving traditional classification and semantic segmentation tasks. The classifier in the fully convolutional networks (FCNs) \cite{long2015fully} can conduct inference and learning on an arbitrary sized image but does not directly output individual object instances. Mask R-CNN \cite{he2017mask} extends Faster R-CNN \cite{ren2015faster} by adding a branch for predicting object masks on top of bounding box extraction.
\cite{dai2016instance_aware,dai2016instance_sensitive} use a multi-task cascaded structure to identify instances with position-sensitive score maps. FCIS \cite{li2016fully} proposed inside and outside maps to preserve the spatial extent of the original image. \cite{bertasius2016semantic} observed that the large receptive fields and the amount of pooling layers of these networks can degrade the quality of instance masks, causing aliasing effect. Region proposal network (RPN) \cite{ren2015faster} can only capture the rough shape of the object and its extensions \cite{bertasius2016semantic,Lin2016RefineNet,pinheiro2016learning} aim to improve the segmentation boundary.

{
\textbf{Single-Image Depth Estimation.}
Classic methods on monocular depth estimation mainly relied on hand-crafted features and graphical models\cite{hoiem2005geometric,saxena2006learning}. More recently, several learning-based approaches boost the performance by utilizing deep models. \cite{eigen2014depth} employed a multi-scale deep network with two stacks for both global and local prediction to achieve depth estimation on a single image. \cite{mousavian2016joint} used CNN for simultaneous depth estimation and semantic segmentation.
However, these works only seek to obtain the relative 3D relationship between different layers and therefore their depth results are much less accurate and clearly insufficient for high quality reconstruction. In our reconstruction, the surface is highly curved but smooth and therefore the depth map needs to be at an ultra-high accuracy, which is extremely difficult to achieve even under the stereo setting, let alone single-image.
}

\textbf{Single-Image 3D Reconstruction.}
Recovering 3D shape from a single image is a long standing problem in computer vision \cite{Brooks:1979}, stemming from image metrology \cite{Reid:1996:GVM,Criminisi:2000:SVM}. The problem is inherently ill-posed and tremendous efforts have focused on imposing constraints such as geometric priors \cite{Debevec:CSD96,Wilczkowiak:2005:UGC}, symmetry \cite{Francois:2002:RMS,Hong2004,Jiang:2009:SAM}, planarity constraints \cite{Zhang:2015:OSA}, shape priors \cite{Saxena:2008:MDP,Huang:2015:SRV}, etc., \youyi{or relying on stock 3D models for 2D-3D alignments \cite{kholgade20143d,aubry2014seeing,rematas2016novel,bansal2016marr}.} Latest approaches leverage deep learning techniques \cite{xiao2012localizing,hejrati2016categorizing,dwibedi2016deep} on large datasets. Eigen \etal \cite{Depth2014} infer depth maps using a multi-scale deep network. The 3D-R2N2 \cite{choy20163d} attempts to recover 3D voxels from a single and multiple photographs. \cite{fan2016point} recovers a dense set of 3D point cloud using a generation network. It is also possible to incorporate 3D geometry proxies such as volumetric abstraction \cite{sala20153,tulsiani2017learning}, hierarchical CSG tree \cite{chen2017progressive}, part models \cite{sala2010contour}, etc. Results from these techniques are promising but still fall short compared with CSG models. Closest to ours is the work from the Magic Leap group, with a clear interest in virtual and augmented reality, to recognize and reconstruct 3D cuboids in a single photograph \cite{dwibedi2016deep}. Our approach is able to recover more general shapes, namely generalized cuboid and cylindrical objects.

\textbf{Sweep-based 3D Modeling.}
A core technique we employ is 3D sweeping. Sweeping a 2D profile along a specific 3D trajectory is a common practice for generating 3D models in computer-aided design (CAD). Early CAD systems \cite{requicha1982solid} use simple linear sweeps (sweeping a 2D polygon along a linear path) to generate solid models. Shiroma \etal \cite{shiroma1991generalized} develop a generalized sweeping method for CSG modeling. Their technique supports curved sweep axis with varying shapes to produce highly complex objects. \cite{angelidis2006swirling} conducts volume preserving stretching while avoiding self-intersections. More recent 3-Sweep \cite{chen20133} and its extension, D-Sweep \cite{Hu2014}, pair sweeping with image snapping. All previous approaches require manual inputs from the user whereas we focus on fully automated shape generation.


\begin{figure*}[t!]
\centering
  \includegraphics[width=\linewidth]{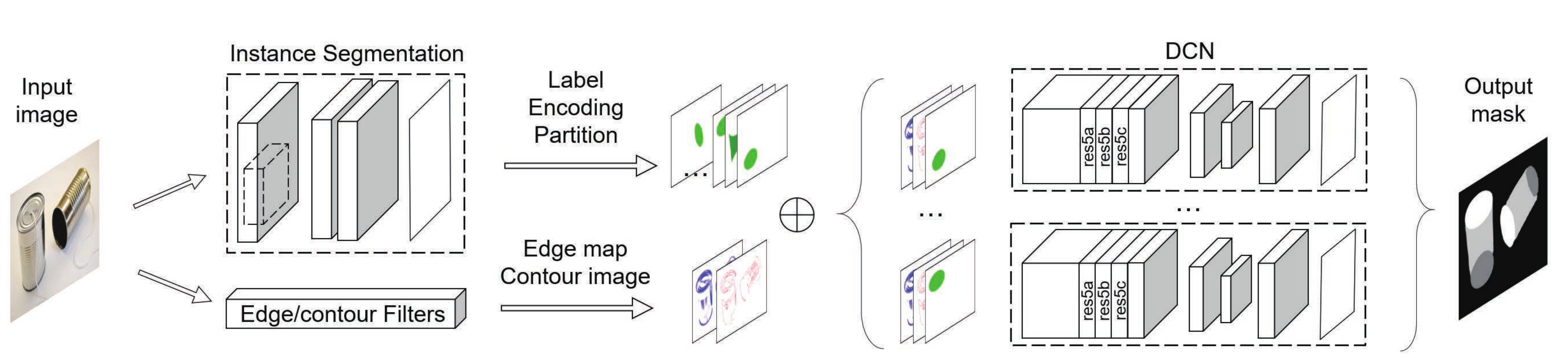}
  \caption{The structure of our GeoNet is composed by an instance segmentation network (Mask R-CNN) and a deformable convolutional network derived from \cite{chen2016deeplab,dai2017deformable}. The net outputs instance masks labeled as semantic parts (profiles, bodies).}\label{fig:network}
\end{figure*}

\section{Overview}

The pipeline of our framework is shown in Fig. \ref{fig:modeling-pipeline}. We take a single photograph containing objects of interests and feed it into our GeoNet to produce instance masks labeled as \emph{cuboid profile}, \emph{cuboid body}, \emph{cylinder profile}, and \emph{cylinder body}. These instance masks are then used for estimating the 3D profile \youyi{(a circle or a rectangle)} and the camera pose, along with a trajectory axis \youyi{(a planar 3D curve)} for the profile to sweep to create the 3D model.

The architecture of our GeoNet is illustrated in Fig. \ref{fig:network}. We build upon the instance segmentation network of Mask R-CNN. The output of Mask R-CNN, coupled with contour image and the edge map, is fed into a deformable convolutional network which is derived from \cite{chen2016deeplab} and \cite{dai2017deformable}. With the information of contour and edge maps, DCN is capable of learning a better and smooth boundary. Details are given in Section \ref{Sec:GeoNet}.

To sweep a primitive part, we first co-relate profile/body masks which could constitute a 3D part. Given correlated profile-body masks, a 3D profile is optimized with camera FoV and a trajectory axis is computed from the body/profile masks. Then, sweeping is performed in 3D to progressively transform and place the estimated 3D profile along the trajectory axis to construct the final model.

\section{Instance Segmentation}

\subsection{GeoNet}\label{Sec:GeoNet}
\begin{figure}[b!]
  \centering
  \includegraphics[width=\linewidth]{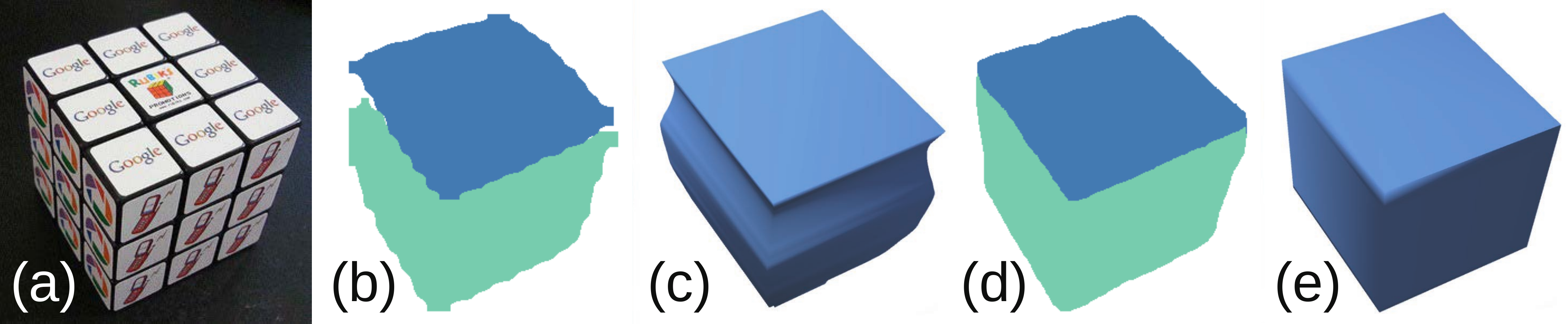}
  \caption{(a) Input image. Segmentation (b) and modeling (c) results of Mask R-CNN. Our GeoNet is capable of filling the gaps and snapping to the boundary (d), (e).}\label{fig:compare_maskrcnn}
\end{figure}

Our GeoNet takes an image as input and outputs the following four types of instance masks: \emph{cuboid profile}, \emph{cuboid body}, \emph{cylinder profile}, and \emph{cylinder body}. A direct instance segmentation network (Mask R-CNN) could lead to erroneous boundaries and incomplete masks that do not agree with the actual object mask, \youyi{because the resolution of feature map are lower due to the ROI memory consumption \cite{he2017mask}.} (Fig. \ref{fig:compare_maskrcnn}). \youyi{Atrous convolution of Deeplab controls the respective fields under a reasonable range, while deformable convolution causes more effective respective fields which can improve the detail of segmentation results.} Thus, we integrate deformable convolution layers proposed in \cite{dai2017deformable} into the network structure of Deeplab \cite{chen2016deeplab} and concatenate it with Mask R-CNN for \youyi{segmentation refinement}. We call the sub network concatenated to Mask R-CNN the deformable convolutional network (DCN). 

To boost the performance of our GeoNet, instead of directly feeding into the DCN with the results from Mask R-CNN, we use more information from the original image to help GeoNet learn more boundary features. { We have tested various case, including using different combination of the original image, the edge map of the original image, and the probability maps from Mask R-CNN, etc., to feed into DCN. Quantitative comparisons are demonstrated in Section \ref{sec:expgeo}. At last, we find combining the edge map \cite{chen20133} and contour map \cite{cheng2009curve} of the input image with probability maps given by Mask R-CNN achieves the best performance. We thus combine these with each instance probability map and feed into DCN. Specifically, for each instance probability maps $I_p^n (n=1,...,N)$ from Mask R-CNN, we combine it with the edge map $I_e$ and the contour map $I_c$ and convert them into a single image ($I_p^n$ takes the Green channel, $I_e$ and $I_c$ take the Red and Blue channel respectively, see Fig. \ref{fig:network} middle). We assign different green values (40 for cuboid body, 100 for cuboid profile, 150 for cylinder body, and 200 for cylinder profile) weighted with probability map $I_p^n$ for different instance categories to distinguish the instances. The shape of instances in one category have quite similar geometrical characteristics, thus labeling the instance with different green values helps the network to learn a better geometrical feature within this category. We find this simple strategy greatly improves the performance of DCN.}

The output of DCN is a refined instance mask $\hat{I_m^k}, k \in \{1,...,N\}$. After getting through the DCN, we combine all instance masks $\hat{I_m^n}$ $(n=1,...,N)$ to form the final mask. To enforce feature learning, the beginning of our DCN is formatted by Res-Net with deformable convolution layers in res-5a, res-5b and res-5c, and connected with 2 convolution layers and 1 deconvolution layer. 





\begin{figure}[t!]
  \centering
  \includegraphics[width=\linewidth]{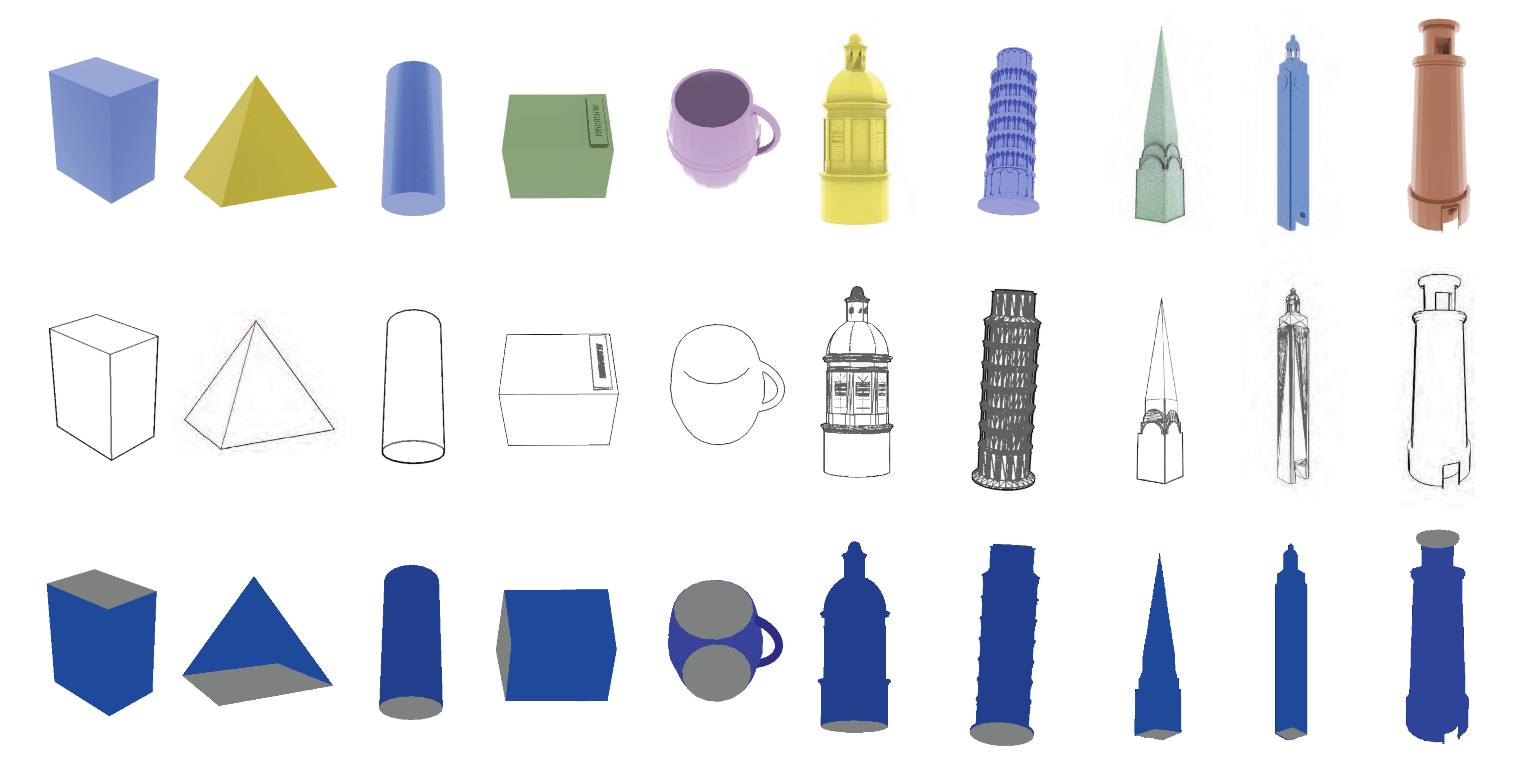}
  \caption{\youyi{Representative synthetic models used in our pre-training. The second and third rows are the corresponding contour maps and label masks, respectively.}}\label{fig:synthdata}
\end{figure}

\textbf{Pre-training.}\label{sec:pre_train} Large nets are typically difficult to train. A good initial guess of the parameters usually leads to better convergence. Thus before using the real images, we pre-train the net with synthetic data. We manually construct a dataset containing 10 exemplar cuboids and generalized cylinders collected from ShapeNet \cite{chang2015shapenet} (see in Fig. \ref{fig:synthdata}). We render these examples from uniformly sampled view angles to generate 1000 images for each example, which gives us 10000 examples for pre-training. Since we do not have a large number of instances in our dataset, we decrease the ROI number from 256 to 128 during the training of Mask R-CNN. We also enlarge our dataset with flipped images.



\section{Modeling}
Given the output masks from GeoNet, our next task is to create a 3D model that agrees with the target masks. We first separate the masks into independent parts (i.e., primitives) constituted by profiles and body and then construct each part independently.

\subsection{Instance labelling}\label{mrf}
Let us denote the set of instances segmented from the network as unlabelled profile faces $\Omega = \{f_1,f_2,...,f_n\}$ and labelled bodies $\Gamma = \{l_1,l_2,...,l_m\}$. Our task is to match each unlabelled profile $f_i$ with its corresponding body $l_j$. This is essentially a labeling problem.

We formulate the following minimization problem:
\begin{equation}\label{e}
  \argmin_{\{{k,l}\}} E := \sum{E_u(f_i \rightarrow {l_k})} + \lambda{\sum{E_b(f_i \rightarrow l_k,f_j \rightarrow l_l)}},
\end{equation}

where $E_u(f_i \rightarrow {l_k}) = \alpha E_1(i,{k}) + (1-\alpha) E_2(i,{k})$ is the unary term. $E_1$ measures the closest Euclidean distance between profile $i$ and body ${k}$. We set it to a large constant $C=1000$ if the distance exceeds a threshold $D$ ( 3\% of the image height in our implementation). $E_2$ measures the proximity of the face to the body. We define it as $E_2(i,{k}) = e^{{-\phi(i,{k})^2}/{2\sigma^2}}$, where $\phi(i,{k})$ is the portion of the points on profile $i$ which are inside the oriented bounding box of body ${k}$. Both $\alpha$ and $\sigma$ are set to 0.3.

The binary term is defined as $E_b(f_i \rightarrow l_k,f_j \rightarrow l_l)=C*\delta(f_i\bigotimes{f_j} | k,l)$, where $\delta(f_i\bigotimes{f_j} | k,l)$ is a function which takes value 1 if $f_i$ and $f_j$ overlaps and $k$ is equal to $l$ and takes value 0 otherwise. The binary term is basically set to penalize two overlapped (i.e., occluded) profiles being assigned to the same body. We solve the above optimization by MRF.

For bodies that have no corresponding profiles, such as the handle of a mug whose profile is invisible due to occlusion, we gather them to form a handle set $\Gamma_H$ and attach them to the closest bodies in $\Gamma$. Fig. \ref{fig:modeling-pipeline} left gives a brief illustration. \youyi{We discard false detected handles if their distance is far away from any detected body ($3\%$ of the image height in our experiments)}.


To fit our 3D model, we use perspective projection rather than orthogonal (which was used in \cite{chen20133}) to create 3D models resembling real world objects. Direct global optimization of the primitive and camera parameters could easily render the problem difficult due to the large variable space. We thus decouple the problem into three steps: profile fitting, trajectory axis estimation, and 3D sweeping.

\subsection{Profile fitting}
As the object profiles in our case are circles and rectangles, this imposes strong priors for our optimization. We assume a fixed camera pose and camera-to-object distance. 
Below are details for fitting the 3D circle and rectangle respectively. The key is to find a plausible initial value for the optimization.


\textbf{Circle.}
Circles in 3D become ellipses in 2D after projection. We use the PCA center $c$ as the initial circle center, with a default depth value $10$. The 3D position of the endpoints $v_1,v_2$ of the PCA major axis are also obtained at depth 10. The initial radius $r$ is then assigned according to the length of the 3D major axis. For the circle orientation, we cast a ray from the camera to one of the endpoints of the minor axis to intersect with the sphere of radius $r$ centered at $c$. Let $s$ be the intersecting point. The orientation is set as the normal of the plane passing through $v_1$, $v_2$, and $s$.

Given the initial circle $C$, together with the mask outline, we optimize 5 variables using Levenberg-Marquardt. The 5 variables are $n=(n_x,n_y,n_z)$, $r$ and $f$ which is the field of view (FoV) of the camera. 
We define the following optimization formulation:
\begin{equation}\label{c}
\argmin E := E_p + E_{f}
\end{equation}
$E_p$ stands for alignment error after projection, it is defined as $E_p := \tau(p | m) + {\alpha/r}$, where $\tau(p | m)$ denotes the portion of points which are not inside the mask. $\alpha$ is set to 40. $E_p$ ensures the circle is inside the mask boundary while its radius is as large as possible after the projection. $E_{f}$ stands for the error between profile normal and the starting direction of the trajectory axis (Section \ref{axis}) under different FoVs. We define it as follows: $E_{f} := \Theta(n_p,n_s) + \eta(n_x,n_y,n_z)$, where $\Theta(n_p,n_s)$ is the acute angle between $n_p$ and $n_s$, with $n_p$ denoting the normal $n$ projected to 2d and $n_s$ denoting the starting direction of the medial axis mentioned in Section \ref{axis}. $\eta(n)$ is a function that guarantees normal $n$ has a square magnitude of 1.

In a second step, we optimize the circle position $c$ separately using only the first term of the objective function $E_p$ to get an updated $c$. With the new $c$, we go back to the optimization of radius, normal and camera FoV. The two steps are iterated until convergence.

\textbf{Rectangle.}
Rectangles are optimized in a similar way. We first detect four vertices by fitting a quadrilateral to the profile mask. Then cast four rays from the camera to the four vertices. The 3D vertices $v_1, v_2, v_3, v_4$ (in clockwise) of the four vertices which lie on the four rays are then optimized as follows:
\begin{equation}\label{ro}
\begin{array}{l}
\argmin E := E_c + E_p + E_{f}
\end{array}
\end{equation}
where $E_c$ keeps the spatial information of the rectangle through the following constraints: (1) parallel edges have equal length, (2) adjacent edges are perpendicular to each other, (3) four vertices are coplanar. We define $E_c$ as:
\begin{equation}\label{cec}
\begin{array}{c}
E_c := \sum_{i=1}^4(\lambda_1(|e_i|-|e_{i+2}|) + \lambda_2\Theta(e_i,e_{i+1}) \\ \\ + \lambda_3\Theta(e_i \times e_{i+1} , e_{i+1} \times e_{i+2})))
\end{array}
\end{equation}
where $e_i$ are the vector created by adjacent vertices $v_i,{v_{i+1}}$. $\Theta$ computes the {cosine of the} acute angle between two vectors. We add parameters $\lambda_i$ to normalize each term. $E_p$ and $E_f$ are the same as above {with radius replaced by side length.} {We rectify the 3D vertices to form a strict planar rectangle during iteration.}




\subsection{Trajectory axis extraction}\label{axis}

\begin{figure}[t!]
\begin{center}
   \includegraphics[width=\linewidth]{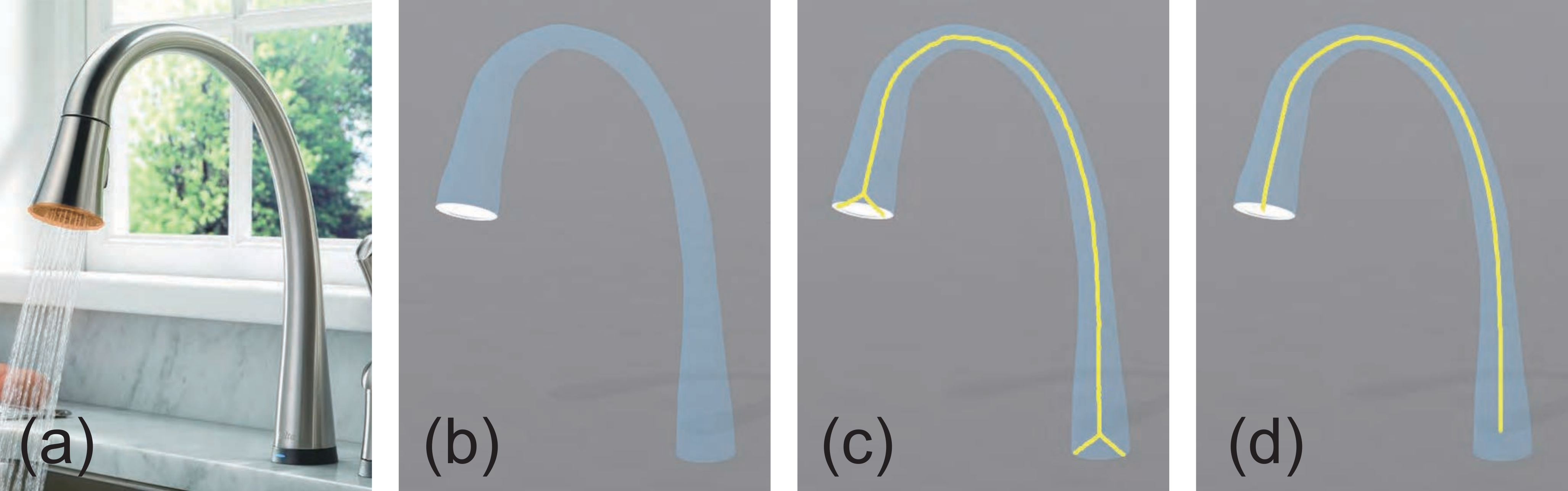}
\end{center}
   \caption{(a) Original image with profile mask. (b) 3D profile (in white). (c) Trajectory axis after thinning. (d) Trajectory axis after pruning.}
   \label{fig:modeling}
\end{figure}

We then extract a trajectory axis that approximates the main axis of the body. The curve will be a guiding line for the sweeping procedure. We use a morphology operation called thinning \cite{lam1992thinning} to get a single width skeleton of the mask image, as shown in Fig. \ref{fig:modeling}, (b). To better account for the completeness of the skeleton, we use both body and profile masks for thinning. To remove the spurious branches in the skeleton, we use a simple way to prune the branches. We mark the skeleton points as branching point and end points using hit-or-miss \cite{Dougherty1992An}. Branches are identified as paths connecting end points and branching points. We progressively delete shortest branches until we get no branching point.

\begin{figure}[t!]
  \centering
  \includegraphics[width=\linewidth]{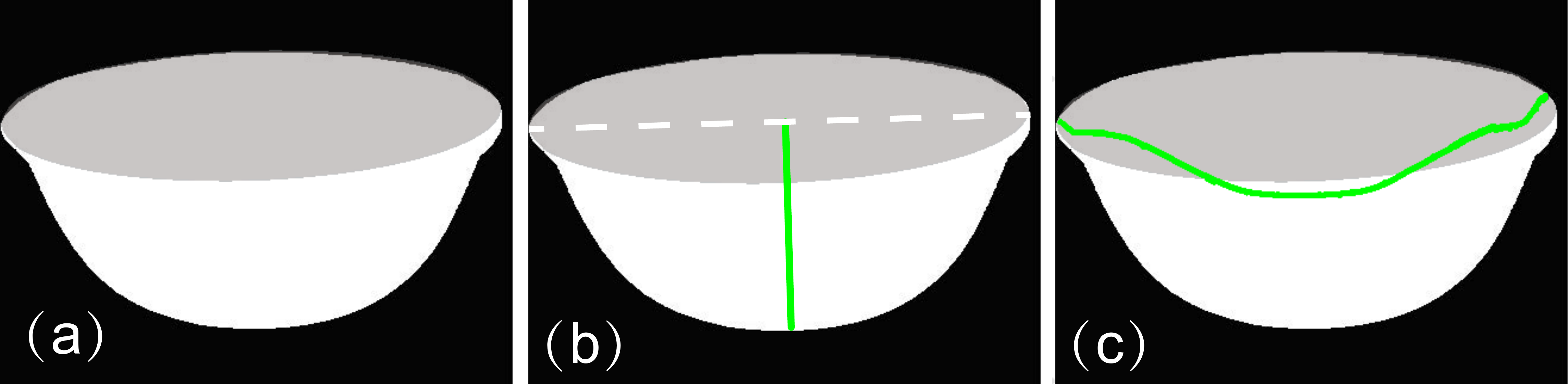}
  \caption{Trajectory axis extraction. (a) The input mask image. (b) Our result. (c) Medial axis extracted using the method of \cite{SOLISMONTERO2012477}}\label{fig:skeleton}
\end{figure}

As our purpose is to reconstruct cylindrical and cuboid object whose trajectory axis is either a straight line or a curve. 
We perform trajectory axis classification. The goal is to classify whether the trajectory axis is a straight line or not. Simple heuristics such as using line fitting with specific thresholds could lead to erroneous estimations. For a more general solution, we utilize the training data available in our dataset. We employ the LeNet \cite{Lecun98gradient-basedlearning} and modify the last FC layer into 2 classes. We use both the body mask and the associated profile masks as input to provide the net with more contextual information. Specifically, we compute their bounding box and scale them to the size of $56\times56$ as input to the net. We get an accuracy of $96\%$ for this task.

If the trajectory axis is labeled as a straight line, we rectify the axis direction w.r.t. profile axis in cases the thinning process gives erroneous skeleton (e.g., for a cylinder we simply set the axis to be orthogonal (in 2D) to the major axis, see Fig. \ref{fig:skeleton}). 
In case when the trajectory axis is labeled as a curve. We set the starting point to profile center and perform bilateral filtering to get the final curve axis. See Fig. \ref{fig:modeling} (d) for an example. \youyi{We find this simple thinning-and-rectifying strategy to perform well in our experiments.}

{We also investigated previous medial axis extraction method of \cite{SOLISMONTERO2012477}. Since their method disregards the context information of the profile faces and thus could lead to erroneous estimations (see an example in Fig. \ref{fig:skeleton}).}


\subsection{Sweeping}
Given the 3D profile and the trajectory axis, our next task is to sweep a 3D model which approximates the body mask. As in \cite{chen20133}, we assume that the trajectory axis lies on a plane which is orthogonal to the profile plane and passes through the profile center. For simplicity, we set the plane orientation to be orthogonal to the camera direction if the object is a generalized cylinder. For a cuboid, we let the plane pass through one of the diagonal lines of the rectangle profile.

We project the 2D body mask and the trajectory axis on that plane and start to place the 3D profile uniformly along the projected trajectory axis. For each part to sweep, we start with the profile with a smaller fitting error if there are two. {For each individual profile $F_t$, $t$ stands for frame index, we cast a 3D ray from its center $c_t$ to intersect with the projected body mask and regard this distance as an initial guess for the profile radius. The final radius $r_t$ of $F_t$ is optimized with}
\begin{equation}\label{radius optimal}
    \min \sum_{k=1}^{n}{M_{\mathbf{Pr}(F_t^k)}} + \sum_{e,s=1,2}(\hat{P}_{e}-P_{s}) + \frac{\alpha}{r_t}
\end{equation}

{Here $F_t$ is the intermediate sweeping profile. $F_t^k \in \mathbb{R}^3 (k=1,2...n)$ represents the sampling points of profile $F_t$.
M is a 2D logical matrix representing the segmentation mask.
$Pr(\cdot)$ is a 3D-to-2D projection function which outputs a 2 dimensional vector $(x,y)$ in the camera space. The vector is regarded as the index of $M$ with $x$ and $y$ representing row and column respectively.
In Eqn \ref{radius optimal}, the first term measures how many sample points fall inside the body mask; the second term is the distance between the
intersection points $\hat{P}_{e}$ and its nearest point $P_{s}$ on the profile boundary as in \cite{chen20133}. $\hat{P}_{e}$ is computed by casting a 2D ray from the projected center $Pr(c_i)$, then intersect with the edges on the edge map $I_e$. Here we reuse the edges of $I_e$ mentioned in Section\ref{Sec:GeoNet}; the third term aims to ensure that the radius is not too small.  $\alpha$ equals 0.025 in our experiment.}

The above procedure optimizes the radius for individual sweeping profiles. To ensure the continuity of the geometry, {we perform a global optimization on all swept profiles $\mathbf{F}$ after the individual frame optimization. For all $\mathbf{T}$ frames, the aim is to refine all centers $\mathbf{C}\in \mathbb{R}^{T\times3}$ and orientations $\mathbf{D}\in \mathbb{R}^{T\times 3}$. 
}
\begin{equation}\label{LPProblem}
      \min \sum\limits_{\mathbf{\Theta} = \mathbf{C}, \mathbf{D}} \ \
      \mathbf{||\Delta(\Theta)||}_2+\mathbf{W||\Theta-\Theta'||}_2,
\end{equation}
$\mathbf{\Delta}$ is the Laplacian smoothing operator. Here the first term in Eqn \ref{LPProblem} measures the smoothness of the geometry, and the second is the deviation of $\mathbf{C}$ and $\mathbf{D}$ to initial values from frames, {every weight inside $\mathbf{W}$ is computed by the dot product between the tangential directions of the current and the next frame center on the trajectory axis.}
Eqn \ref{radius optimal} and \ref{LPProblem} are iterated to get the final result. \youyi{In our experiments, both optimizations take around 1-3 iterations to converge.}

For generalized cylinder or cuboid which have no associated profiles (e.g., a teapot handle), we estimate an initial position and radius for the profile by analyzing the contact region to the part of the already constructed 3D body. The sweeping process is performed similarly to finally create those parts (see Fig. \ref{fig:modeling-pipeline}, \ref{fig:results}). Note that before the sweeping process, we globally optimize the camera pose (FoV) with all estimated 3D profiles.



\begin{table*}[t!]
\centering
\begin{tabular}{l|c|c|c|c|c|c|c|c|c|c}
\hline
    Method              & cub & cuf & cyb & cyf & mAP@0.7  & cub & cuf & cyb & cyf & mAP@0.9   \\ \hline \hline
    FCIS                    &68.19&61.24&50.33&37.51&54.32&33.04&23.71&10.51&9.09&19.09\\ \hline
    GeoNet w. FCIS          &68.61&\textbf{61.47}&56.75&37.23&56.01&48.64&36.88&\textbf{17.14}&10.30&28.24\\ \hline

    Mask R-CNN              &68.36&61.22&55.93&\textbf{40.26}&56.44&35.73&30.13&7.29&10.17&20.83\\ \hline
    GeoNet w. Mask R-CNN    &\textbf{69.49}&61.04&\textbf{57.90}&37.84&\textbf{56.57}&\textbf{50.18}&\textbf{37.92}&13.89&\textbf{11.37}&\textbf{28.34} \\ \hline


\end{tabular}
\caption{Evaluation of GeoNet with FCIS \cite{li2016fully} and Mask R-CNN \cite{he2017mask} at overlap thresholds of 0.7 and 0.9 respectively.}
\label{netcomp}
\end{table*}

\section{Experiments}
\textbf{Dataset.}\label{dataset}
Besides the synthetic data described in Section \ref{sec:pre_train}, our real dataset contains multiple human-made primitive-shaped objects widely used in daily life such as mugs, bottles, taps, cages, books, and fridges, etc. There are 11657 real images and 10000 synthetic images (with 11590 generalized cuboids and 15008 generalized cylinders). \youyi{The real dataset contains about $6000$ unannotated images from ImageNet \cite {deng2009imagenet}, $774$ annotated images from Xiao et al. \cite{xiao2012localizing}, and $4883$ images collected from the Internet.} 
The real dataset is further separated into $8183$ training images and $3474$ testing images. We perform evaluations of all experiments on the testing set of real images.

\textbf{Experiment of GeoNet.}\label{sec:expgeo}
{In order to make full use of the information from original image as well as the outputs of instance segmentation network, We test various combination of gray map $I_g$, edge map $I_e$, contour map $I_c$ of the image, mask $I_m$, probability map $I_p$ from the network.
We restrict the combination to form a three channel image, and duplicate channels when the assembled map number is less than 3.
For this experiment of combination strategy, we adopt Mask R-CNN as the first stage of our GeoNet. We use mean intersection-over-union (mIoU) defined over image pixels as the evaluation metric, since we are focusing on boundary refinement because the instances are the same during these experiments.
The results are shown in Table \ref{bnfeval}, the combination of $I_c, I_p, I_e$ significantly outperforms the others.}

\begin{table}[h!]
\centering
\begin{tabular}{l|c|c|c|c|c} \hline
    Method  & cub & cuf & cyb & cyf & mean   \\ \hline \hline
    Mask R-CNN                       &77.56 & 80.51 & 68.68 & 75.74 & 75.62 \\ \hline
    GeoNet w. $I_m$                  &87.51 & 85.50 & 77.89 & 82.87 & 83.44     \\ \hline
    GeoNet w. $I_p$                  &89.34 & 85.84 & 79.01 & 83.19 & 84.34              \\ \hline
    GeoNet w. $I_g,I_p$ 	       	 &90.12 & 85.92 & 78.28 & 83.22 & 84.39   \\ \hline
    GeoNet w. $I_e,I_m$              &89.67 & 86.03 & 79.78 & 83.82 & 84.83     \\ \hline
    GeoNet w. $I_e,I_p$              &90.88 & \textbf{86.84} & 79.51 & 84.36 & 85.40              \\ \hline
    GeoNet w. $I_c,I_m,I_e$          &91.80 &86.24 & \textbf{85.27} & 85.37 & 87.17 \\ \hline
    GeoNet w. $I_c,I_p,I_e$   	 &\textbf{92.47} & 86.81 & 84.72 & \textbf{87.02} & \textbf{87.76} \\ \hline

\end{tabular}
\caption{Evaluation of GeoNet on different combinations of gray map $I_g$, edge map $I_e$, contour map $I_c$ of image, mask $I_m$, probability map from Mask R-CNN.}
\label{bnfeval}
\end{table}

Since our GeoNet is built upon existing instance segmentation networks, to evaluate its effectiveness, we experimented with {generally accepted} networks of FCIS \cite{li2016fully} and Mask R-CNN \cite{he2017mask}. We attach the DCN to both FCIS and Mask R-CNN and evaluate the performance of improvements in the segmentation results.

\begin{figure}[h!]
  \centering
  \includegraphics[width=\linewidth]{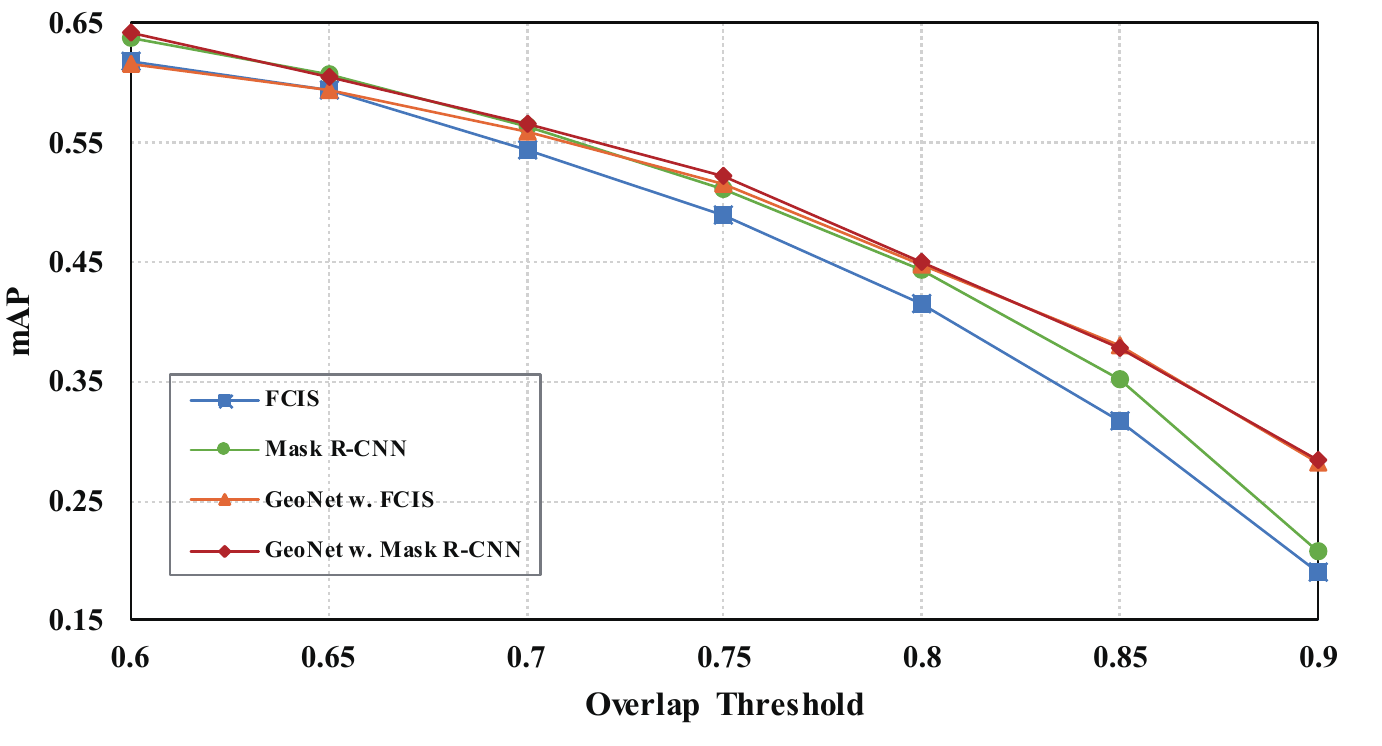}
  \caption{DCN improves the performance of segmentation results when the base segmentation results are more faithful to the ground truth.}\label{netcurve}
\end{figure}

Accuracy is evaluated by mean average precision, mAP\cite{hariharan2014simultaneous}, at mask-level IOU (intersection-over-union) with overlap threshold set to 0.7 and 0.9 respectively.
The results are shown in Table \ref{netcomp}. DCN performs better at larger overlap thresholds. At threshold 0.9, DCN improves the performance by $9.15\%$ and $7.51\%$ (mAP), respectively, which shows that DCN is capable of refining the segmentation result on an adequate basis (see also Fig. \ref{fig:compare_maskrcnn} for a visual comparison). For a plausible comparison, we set the instance count to a fixed number for computing mAP. The chart in Fig. \ref{netcurve} shows the mAP at different overlap thresholds. DCN works better when the base results from FCIS and Mask R-CNN agree with the ground truth. {We only visualize the range [0.6, 0.9] since DCN is capable of boosting the performance when the segmentation results are rather accurate w.r.t. the ground truth, while when mAP is lower than 0.6, we find that DCN is much less helpful for refining the boundary.}

\begin{figure*}[t!]
  \centering
  \includegraphics[width=\linewidth]{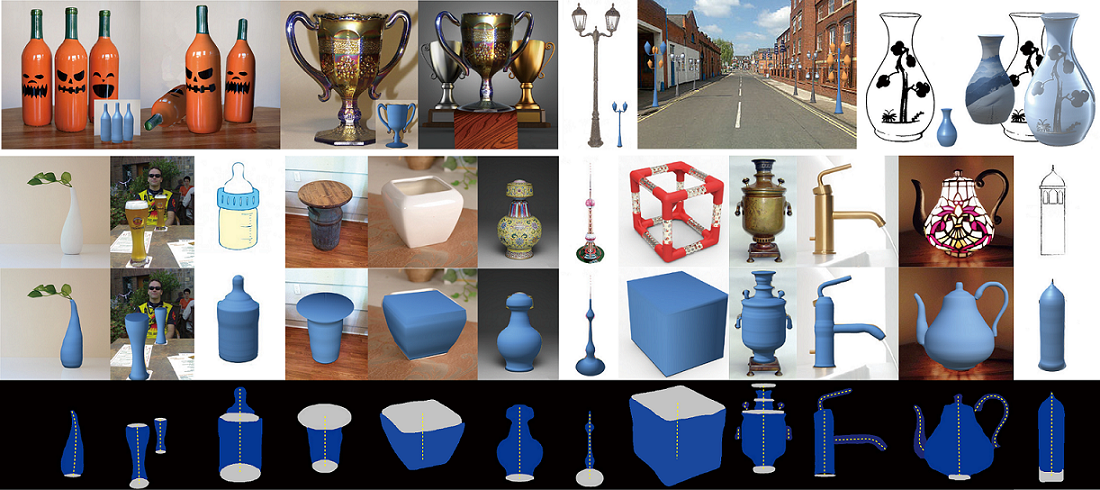}
  \caption{Representative results generated using our method. Our method is able to recover objects constituted by multiple semantic parts (e.g., teapots, lamps, water taps, etc.). The first row shows some of the editing results of the model created. The two examples (last column) show that our method can be directly applied to sketch input. {We assume symmetry in texture maps, mirror the front texture to back, and finally stitch them together.}}\label{fig:results}
\end{figure*}

It is also noteworthy that our method is capable of segmenting and reconstructing objects from raw sketch inputs as shown in the last column of Fig. \ref{fig:results}. This indicates that our DCN network is able to learn cues from the input contour images and edge maps for predicting the final mask.

\begin{table}[h!]
\centering
\begin{tabular}{c|c|c|c|c|c|c}
\hline
    Metric & Method  & cub & cuf & cyb & cyf & mean   \\ \hline \hline
    \multirow{3}{*}{PP-IOU}
    & Baseline                          &  80.97    &  76.66    &  78.75           &  59.85                        & 74.06 \\ \cline{2-7}
    & BNF \cite{bertasius2016semantic}    & \textbf{83.40}    &  \textbf{77.86}    &  79.02           &  58.46     & 74.69 \\ \cline{2-7}
    & Ours                                &  82.94    &  77.69    &  \textbf{80.70}  &  \textbf{60.62}             & \textbf{75.49} \\ \hline
    \multirow{3}{*}{PI-IOU}
    & Baseline                          &  79.50    & 78.52     &  77.70           &  59.36                                  &73.77       \\ \cline{2-7}
    & BNF \cite{bertasius2016semantic}    &  80.39    & 76.67     &  77.19           &  47.81                                &70.52       \\ \cline{2-7}
    & Ours                                &  \textbf{81.47}   & \textbf{79.94}    &  \textbf{78.51}        &  \textbf{59.42} &\textbf{74.84}       \\ \hline
\end{tabular}
\caption{Semantic segmentation comparison on our dataset. Note that BNF has a significant drop on cylinder profile because it may fail when the boundaries are not clear, while many cylinder profiles have no clear boundaries due to self occlusion in our case.}
\label{bnfeval}
\end{table}


\textbf{Comparisons to boundary refinement method.} We compare GeoNet with Boundary Neural Fields \cite{bertasius2016semantic} on semantic segmentation task {on our test set containing 1614 cuboids and 1840 cylinders.} We use the evaluation metrics pixel intersection-over-union averaged per pixels (PP-IOU) and pixel intersection-over-union averaged per image (PI-IOU) same as \cite{bertasius2016semantic}. We also run the evaluation on the Mask R-CNN output as a baseline for the comparison.

According to this metric, PP-IOU is computed on a per pixel basis. As a result, the images that contain large object regions are given more importance. On the other hand, PI-IOU gives equal weight to each of the images. As shown in Table \ref{bnfeval}, BNF has lower accuracy on PI-IOU indicates that it is not able to segment small objects accurately. 
{However our method outperforms Mask R-CNN and BNF on average accuracy on both metrics.}

\textbf{Comparisons to cuboid detection and reconstruction methods.} We use the SUN primitive dataset \cite{xiao2012localizing} to evaluate our method on cuboid reconstruction and compare with the methods of \cite{xiao2012localizing} and \cite{dwibedi2016deep}. For cuboid detection, a bounding box is correct if the Intersection over Union (IOU) overlap is greater than 0.5. For keypoint localization, we use re-projection accuracy (RA) used in a baseline approach Xiao \etal \cite{xiao2012localizing} as well as the Probability of Correct Keypoint (PCK) and Average Precision of Keypoint (APK) metrics used in the state-of-the-art method Dwibedi \etal \cite{dwibedi2016deep}. The latter two are commonly used in the human pose estimation task. We use the re-projection corners of the reconstructed cuboids as keypoints for this task. The comparison results are shown in Table \ref{cuboideval}. 
The numbers show that our approach performs better in both tasks.

\begin{table}[t!]
\centering
\begin{tabular}{c|c|c|c|c}
\hline
    Method  & AP & RA & APK & PCK \\ \hline \hline
    Xiao \etal \cite{xiao2012localizing}  & 24.00 & 38.00 & - & - \\ \hline
    Dwibedi \etal \cite{dwibedi2016deep}   &  75.47 & - &  41.21   & 38.27    \\ \hline
    Ours    &   \textbf{ 79.56 }  &  \textbf{ 49.79} & \textbf{47.56} & \textbf{45.11} \\ \hline
\end{tabular}
\caption{Comparison of cuboid bounding box detection and keypoint localization. AP is the average precision for bounding box detection used in Xiao \etal \cite{xiao2012localizing}. 
}
\label{cuboideval}
\end{table}


\begin{figure}[h!]
  \centering
  \includegraphics[width=\linewidth]{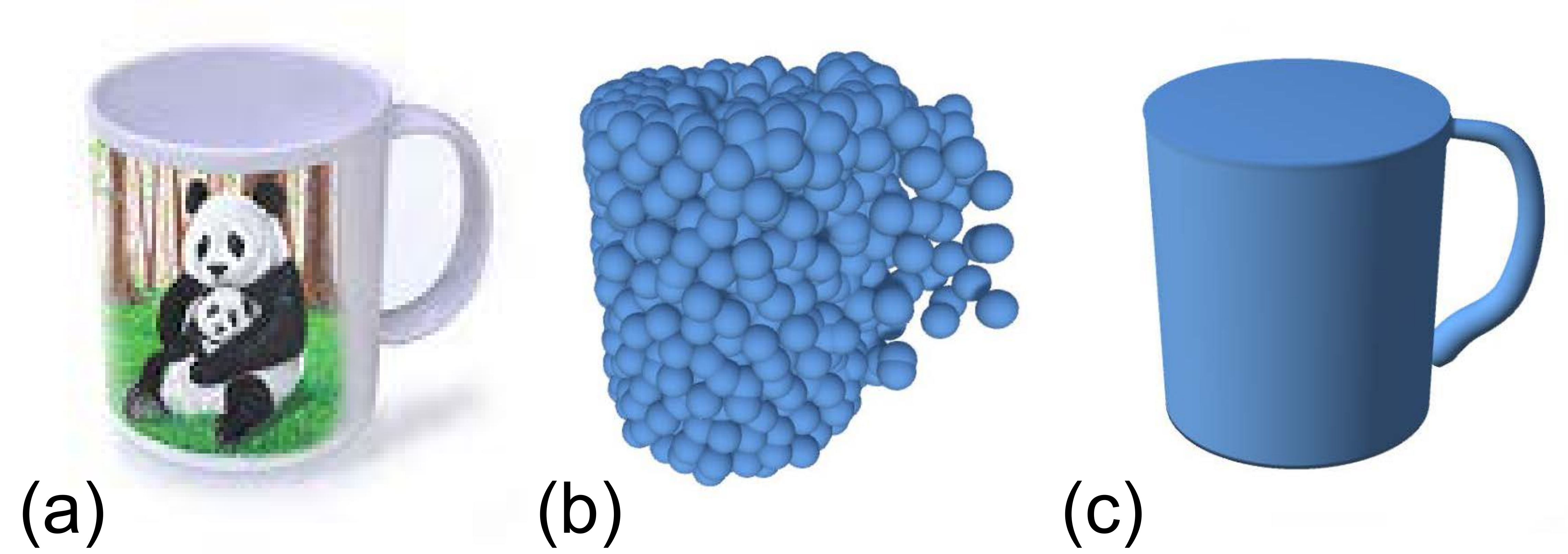}
  \caption{Comparison with point/voxel-based image reconstruction methods. (a) The input image. (b) The result of point-based framework \cite{fan2016point}. (c) Our result.}\label{voxel}
\end{figure}

\begin{figure}[h!]
  \centering
  \includegraphics[width=\linewidth]{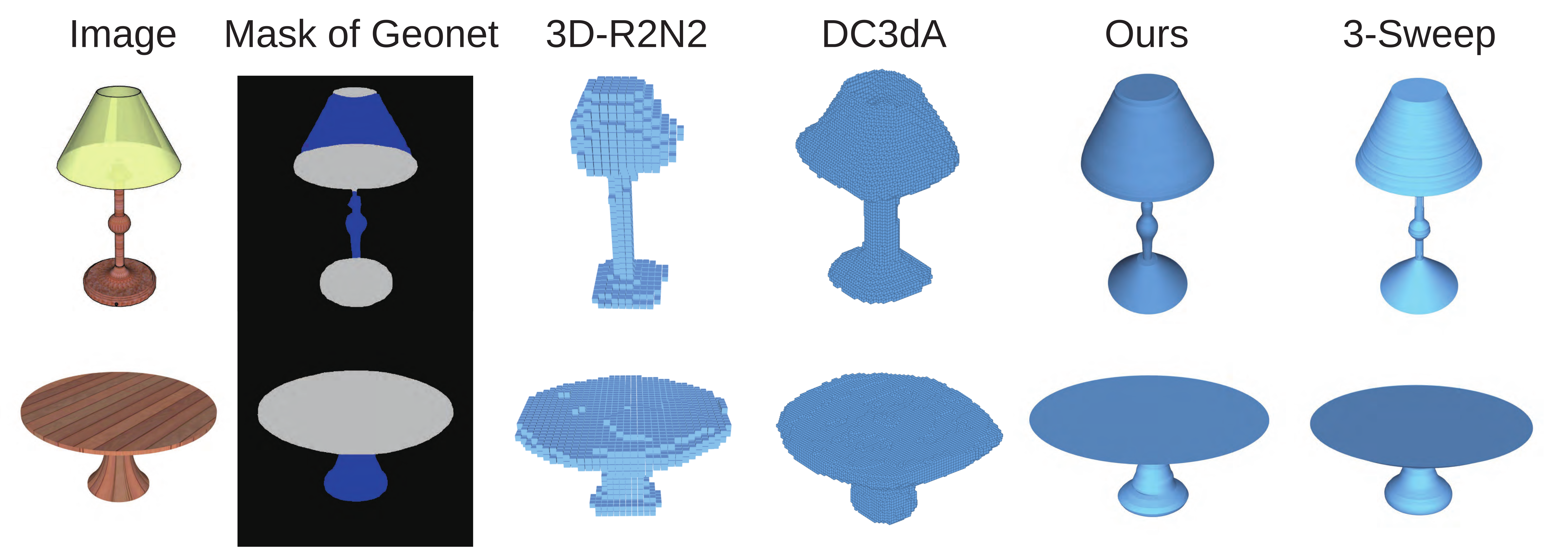}
  \caption{The comparison with 3-sweep, 3D-R2N2 and \youyi{the Densely Connected 3D Autoencoder in Li \etal \cite{yi2017large}.}}\label{semi}
\end{figure}

{
\textbf{Comparisons to point/voxel-based and semi-automatic reconstruction methods.}} We compare our method with two single image reconstruction methods using neural networks, Choy \etal \cite{choy20163d} and Fan \etal \cite{fan2016point}. \youyi{We also compare with Densely
Connected 3D Autoencoder in Li \etal \cite{yi2017large} from the ShapeNet reconstruction challenge. All }of them are able to generate a rough representation of the 3D object from a single photograph.
The visual comparison examples are shown in Fig. \ref{voxel} and {Fig. \ref{semi}. {We train their network using the 2000 cup and 2000 lamp models collected from ShapeNet\cite{chang2015shapenet}.}} The models are generated with the code provided by the authors with default parameters. It can be seen that our result is cleaner and more accurate. In addition, our models can be directly textured and edited while theirs can not due to the lack of semantic part information.

{
Additionally, we conduct experiments to compare our approach vs. semi-automatic method 3-sweep\cite{chen20133} on $10$ models (5 tables, 5 lamps) using our own implementation. The average reconstruction error for 3-sweep is 1.263\% whereas our is 1.262\%. Fig. \ref{semi} shows the qualitative examples.
}

\textbf{Timing.}
The training of the networks is performed on a server with $4$ NVIDIA GeForce GTX Titan X GPUs, an Intel i7-6700K CPU, and 64GB RAM. It takes three days to train the Mask R-CNN and one day to train the DCN on our dataset of 8183 images. {It takes $1s$ for GeoNet to segment one image and less than 1 second to reconstruct objects from the masks including stages of instance labeling, profile fitting, and 3D sweeping with multi-thread acceleration (the individual profile optimization can be performed in parallel).}

\textbf{Limitations.}
\begin{figure}[t!]
  \centering
  \includegraphics[width=\linewidth]{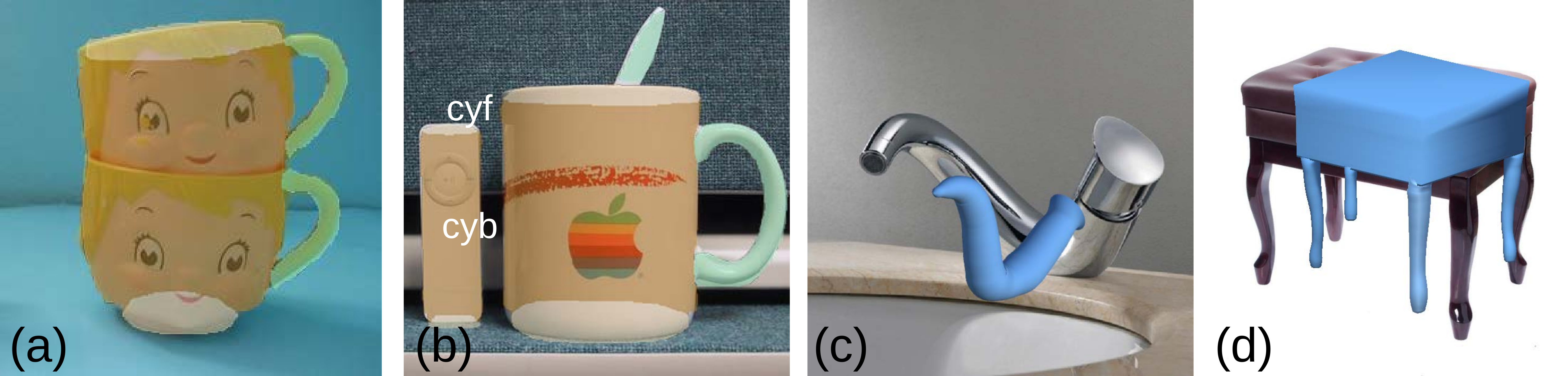}
  \caption{The failure cases of our approach.}\label{fail}
\end{figure}
Our method has a few limitations. As shown in Fig. \ref{fail}, the network is not able to infer the regions of instances which are cluttered or under occlusion. Priors such as symmetry and physical validity can be enforced to alleviate the problem as in \cite{Guo:2013:SSP,Shao:2014:IUS}. Next, the network may also give wrong class labels when the 2D projection of the shape is vague. As shown in Fig. \ref{fail}, the remote control is mistaken for a generalized cylinder by the network. \youyi{For complex objects, our method is currently not able to predict parts which deviate much from the training set or cannot be approximated by GC-GCs such as the parts of the table shown in Fig. \ref{fail}. In this example, it should also be noted that our method may fail to predict correct alignments between the parts. This is because in our experiments,} individual parts are constructed in parallel whereas their semantic relations such as coplanar or co-axial may need further rectification utilizing methods of e.g., \cite{Li:2011:GCF}. In the future, it would be interesting to incorporate such semantics in the network design. Finally, our method cannot handle cases where the axis of the object does not lie on a spatial plane. Thus the object can not have spiral axis such as a spring. To infer such spatially varying curved trajectory requires additional assumptions \cite{Bae:2008:IAS}. We also leave this for future work. 
\section{Conclusion}
This paper presents a fully automatic method for extracting 3D editable objects from a single photograph. Our framework uses Mask R-CNN as a basis to build a network which is capable of improving the instance segmentation results. In the subsequent modeling stage, we simultaneously optimize for the camera pose and the 3D object profile and estimate the 3D body shape by a sweeping algorithm.

Our framework is capable of reconstructing primitive objects constituted by generalized cuboids and generalized cylinders. Unlike previous 3D reconstruction methods which reconstruct either 3D point clouds, voxels, or surface meshes, our model recovers high-quality semantic parts and their relations, which naturally enables plausible edits of the image objects. Qualitative and quantitative results have demonstrated the effectiveness of our method. \youyi{In the future, we plan to explore possibilities of building a more generic and end-to-end framework to reconstruct high-quality primitive 3D shapes from single images or videos.} 


%

\appendices


\ifCLASSOPTIONcompsoc
  \section*{Acknowledgments}
\else
  \section*{Acknowledgment}
\fi

The authors would like to thank all the reviewers for their insightful comments. This work was supported in part the National Natural Science Foundation of China No. 61502306, No. U1609215, the National Key Research \& Development Program of China (2016YFB1001403), and the China Young 1000 Talents Program.

\ifCLASSOPTIONcaptionsoff
  \newpage
\fi



%

\bibliographystyle{IEEEtran}
\bibliography{egbib}


\begin{IEEEbiography}[{\includegraphics[width=1in,height=1.25in,clip,keepaspectratio]{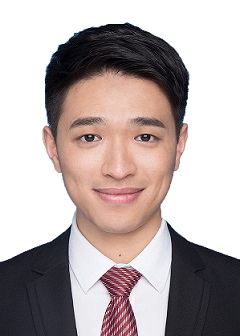}}]{Xin Chen} is a Ph.D. student at the School of Information Science and Technology(SIST), ShanghaiTech University. He obtained his B.S. from the School of Science at Qingdao University of Technology . His research interests include human reconstruction, image-based modeling and deep learning.
\end{IEEEbiography}

\begin{IEEEbiography}[{\includegraphics[width=1in,height=1.25in,clip,keepaspectratio]{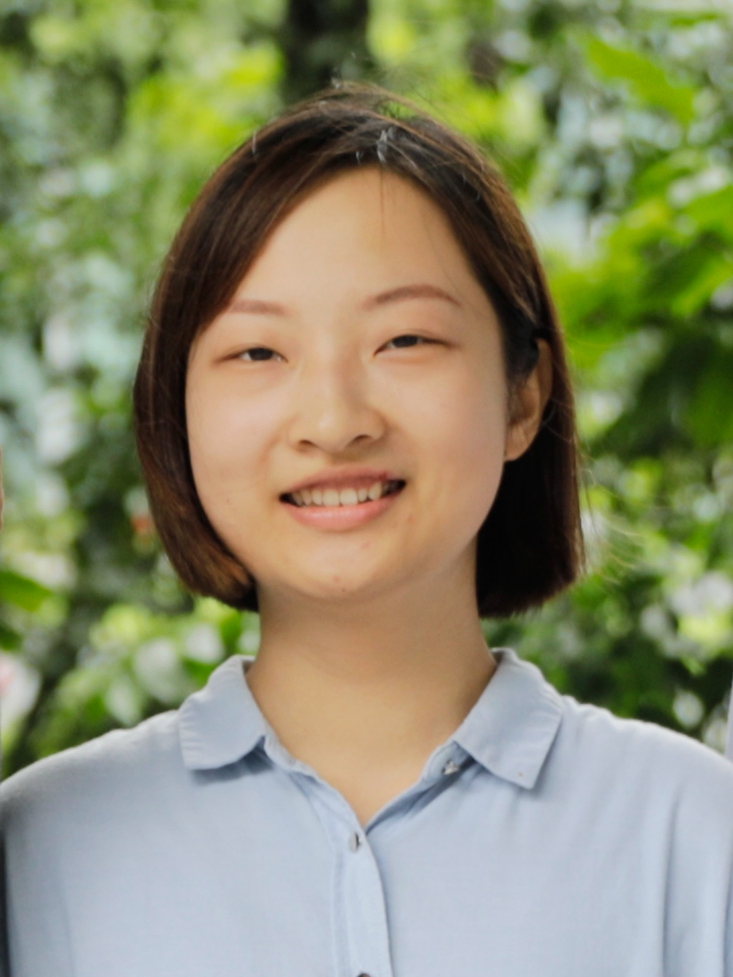}}]{Yuwei Li} is a Ph.D. student at the School of Information Science and Technology(SIST), ShanghaiTech University. She obtained her B.E. from the School of Computer Engineering and Science at Shanghai University. Her research interests include 3D reconstruction, deep learning, and human-computer interaction.
\end{IEEEbiography}

\begin{IEEEbiography}[{\includegraphics[width=1in,height=1.25in,clip,keepaspectratio]{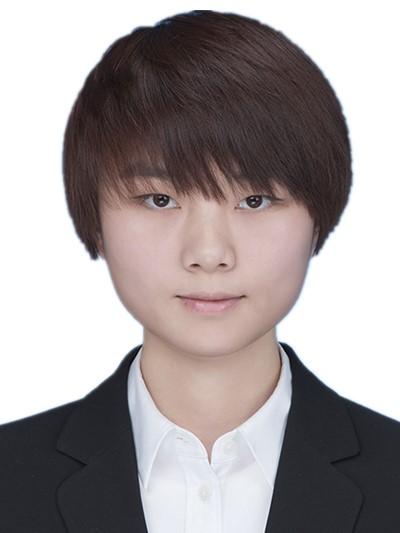}}]{Xi Luo} received her B.S. degree in Communication Engineering from School of Mechanical and Electrical and Information Engineering, Shandong University, Weihai, China, in
2016. Now she is a Ph.D. student in computer science, affiliated with ShanghaiTech University,
Shanghai, China. Her current research interests are 3D reconstruction, virtual fitting, and
human-computer interaction.
\end{IEEEbiography}

\begin{IEEEbiography}[{\includegraphics[width=1in,height=1.25in,clip,keepaspectratio]{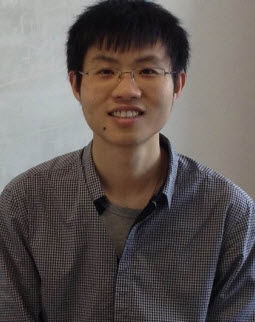}}]{Tianjia Shao} is currently an Assistant Professor at the School of Computing, University of Leeds. Before that, he was an assistant researcher in the State Key Lab of CAD\&CG, Zhejiang University. He received his PhD in Computer Science from Institute for Advanced Study, and his B.S. from the Department of Automation, both in Tsinghua University. His research interests include RGBD image processing, indoor scene modeling, structure analysis and 3D model retrieval.
\end{IEEEbiography}

\begin{IEEEbiography}[{\includegraphics[width=1in,height=1.25in,clip,keepaspectratio]{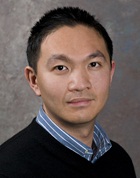}}]{Jingyi Yu} is a Professor in the School of Information Science and Technology, ShanghaiTech University. Before that, he was a full Professor at the Department of Computer and Information Sciences at the University of Delaware. His research covers a range of areas in computer vision including sensors and early vision, computational photography, image-based modeling and rendering, illumination and reflectance, stereo matching, shape-from-X, saliency and segmentation, motion tracking, and medical image analysis. His research has been supported by the NSF, NIH, and DoD. He was a recipient of the NSF CAREER Award in 2009, the Air Force Young Investigator Award in 2010, and the UD College of Engineering Outstanding Junior Faculty Award in 2013.
\end{IEEEbiography}

\begin{IEEEbiography}[{\includegraphics[width=1in,height=1.25in,clip,keepaspectratio]{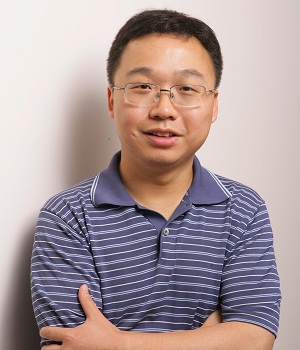}}]{Kun Zhou} is a Cheung Kong Professor in the Computer Science Department of Zhejiang University, and the Director of the State Key Lab of CAD\&CG. Prior to joining Zhejiang University in 2008, Dr. Zhou was a Leader Researcher of
the Internet Graphics Group at Microsoft Research Asia. He received his B.S. degree and Ph.D. degree in computer science from Zhejiang University in 1997 and 2002, respectively. His research interests are in visual computing, parallel computing, human computer interaction, and virtual reality. He currently serves on the editorial\/advisory boards of ACM Transactions on Graphics and IEEE Spectrum. He is a Fellow of IEEE.
\end{IEEEbiography}

\begin{IEEEbiography}[{\includegraphics[width=1in,height=1.25in,clip,keepaspectratio]{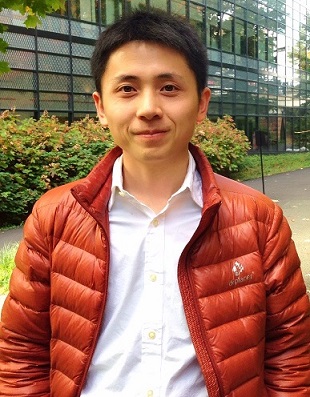}}]{Youyi Zheng} is a Researcher (PI) at the State Key Lab of CAD\&CG, College of Computer Science, Zhejiang University. He obtained his PhD from the Department of Computer Science and Engineering at Hong Kong University of Science \& Technology, and his M.Sc. and B.Sc. degrees in Mathematics, both from Zhejiang University. His research interests include geometric modeling, imaging, and human-computer interaction.
\end{IEEEbiography}

%








\end{document}